\documentclass[a4paper, 10pt, conference]{ieeeconf}  

\usepackage{hyperref}
\usepackage{flushend}
\usepackage{graphicx}
\usepackage{siunitx}
\usepackage{hyperref}
\usepackage{xcolor}
\usepackage{subfig}

\IEEEoverridecommandlockouts                              

\title{\LARGE \bf
A Flexible Field-Based Policy Learning Framework for Diverse Robotic Systems and Sensors}
\author{Jos\'e G. Buenaventura-Carre\'on*$^{1}$, Floris Erich$^{1}$, Roman Mykhailyshyn$^{1}$, \\Tomohiro Motoda$^{1}$, Ryo Hanai$^{1}$ and Yukiyasu Domae$^{1}$
\thanks{$^{*}$Corresponding author, reachable at buenaventuracarreon-gustavo@aist.go.jp}
\thanks{$^{1}$Jos\'e Gustavo Buenaventura-Carre\'on, Floris Erich, Roman Mykhailyshyn, Tomohiro Motoda, Ryo Hanai and Yukiyasu Domae are with the National Institute of Advanced Industrial Science and Technology (AIST), Japan.}%
}

\begin{document}

\maketitle
\thispagestyle{empty}
\pagestyle{empty}

\begin{abstract}
We present a cross-robot visuomotor learning framework that integrates diffusion policy–based control with 3D semantic scene representations from D³Fields to enable category-level generalization in manipulation. Its modular design supports diverse robot–camera configurations, including UR5 arms with Microsoft Azure Kinect arrays and bimanual manipulators with Intel RealSense sensors, through a low-latency control stack and intuitive teleoperation. A unified configuration layer enables seamless switching between setups for flexible data collection, training, and evaluation. In a grasp-and-lift block task, the framework achieved an 80\% success rate after only 100 demonstration episodes, demonstrating robust skill transfer between platforms and sensing modalities. This design paves the way for scalable real-world studies in cross-robotic generalization.
\end{abstract}

\section{Introduction}

Imagine a robot tasked with tidying up a cluttered desk, moving cups, pens, and small objects with the dexterity and adaptability of a human hand. While recent advances in robotic manipulation have enabled impressive performance in structured environments, generalizing these skills to new objects, scenes, and hardware remains a significant challenge. Many existing frameworks rely on fixed robot–camera configurations and task-specific training, limiting their applicability beyond their original setup.

Diffusion policy–based visuomotor learning, combined with rich 3D semantic representations such as D³Fields, has shown that robots can learn flexible manipulation strategies from limited demonstrations. However, these approaches have largely been constrained to specific platforms, for example the ALOHA manipulator with Intel RealSense cameras, hindering broader adoption and comparative studies across different hardware.

In this work, we extend a diffusion policy–based visuomotor learning framework to create a cross-robot system capable of operating across multiple robot–camera configurations. Specifically, we adapt the framework to a UR5 robotic arm with four Azure Kinect cameras, redesigning control, perception, and data acquisition modules while preserving compatibility with the original ALOHA + RealSense setup. This modular design enables seamless switching between hardware setups, supports robust policy training, and facilitates systematic evaluation across platforms. By enabling a single policy framework to operate in diverse physical setups, our work takes a step toward scalable, cross-robot learning, opening the door to broader adoption, reproducibility, and integration with emerging sensor modalities.

\section{Related work}

Recent advances in robot learning have focused on developing generalizable visuomotor policies, improving perception in cluttered environments, and enabling adaptive manipulation across diverse robotic platforms. Approaches such as learning from demonstrations~\cite{oh2025robustinstantpolicyleveraging,10711321} and reinforcement learning~\cite{motoda2025learningbimanualmanipulationaction} have been shown to improve performance in both prehensile and non-prehensile manipulation tasks. Integrating foundation models for enhanced scene understanding has further facilitated robust perception and goal specification~\cite{10871055,10871084}. These works collectively highlight progress in perception, control, and learning strategies for real-world robotic manipulation.

\subsection{D3Fields}

Scene representation plays a pivotal role in robotic manipulation systems. Traditional methods often capture geometric shape but neglect semantic understanding or dynamic scene evolution. D³Fields~\cite{wang2024d3fields} innovates by providing dynamic, semantic, and implicit 3D descriptor fields, an integrated representation that maps arbitrary 3D points to semantic features and instance masks, while capturing environmental dynamics.

D³Fields project 3D points into multiple RGB-D views, extract features using foundational vision models like Grounding-DINO~\cite{10.1007/978-3-031-72970-6_3}, SAM~\cite{Kirillov_2023_ICCV}, XMem~\cite{cheng2022xmem}, and DINOv2~\cite{oquab:hal-04376640}, and fuse them into a descriptor field without any task-specific training. This enables zero-shot generalization to new rearrangement tasks by specifying goals via diverse 2D images.

D³Fields builds upon the foundation laid by F3RM~\cite{shen2023F3RM}, which introduced the idea of using multi-view 2D foundation models to create a unified 3D feature representation without task-specific training. F3RM demonstrated that 3D fusion of features extracted from RGB(-D) views enables effective few-shot transfer in robotic manipulation. D³Fields extends this approach from static scenes to dynamic environments, augmenting the representation with temporal consistency and motion-aware features, thereby broadening its applicability to rearrangement tasks that involve both semantic understanding and scene evolution.

Across both real-world and simulated household manipulation tasks, such as shoe organization, debris collection, and office desk cleanup, D³Fields outperform state-of-the-art implicit 3D representations (e.g.\ Dense Object Nets~\cite{florencemanuelli2018dense}, DINO~\cite{caron2021emerging}) in terms of manipulation accuracy and generalizability.

While various neural field representations have been explored in robotics (e.g.\ Neural Descriptor Fields~\cite{9812146}, Neural Radiance Fields~\cite{10.1145/3503250}), few manage to integrate 3D geometry, semantics, and scene dynamics simultaneously. D³Fields occupy a unique position in this landscape by combining all three in a unified, zero-shot capable framework.

\subsection{GenDP}

Recent advances in diffusion-based control policies have demonstrated substantial abilities to handle complex robotic manipulation tasks. However, their capacity to generalize to unseen objects or layouts is often limited by insufficient modeling of geometry and semantics.

GenDP~\cite{wang2023gendp} presents an imitation learning framework that addresses these limitations by embedding explicit spatial and semantic cues through 3D semantic fields. Their method constructs high-dimensional 3D descriptor fields from multi-view RGB-D data using large foundational vision models; these are then aligned with reference descriptors to generate semantic fields. By fusing these features with raw point clouds and feeding them into PointNet++~\cite{NIPS2017_d8bf84be} and the diffusion policy, GenDP achieves remarkable generalization: on unseen object instances, success rates improve dramatically, from 20\%  to 93\%, across eight varied manipulation tasks involving articulated objects and diverse appearances.

While diffusion-based policies are a powerful paradigm, earlier works typically rely on visual or geometric states without explicitly encoding fine-grained semantics. GenDP stands out by systematically incorporating semantic field representations to resolve geometric ambiguities and capture subtle semantic distinctions, enabling strong category-level generalization.

\subsection{Diffusion Policy}

Diffusion Policy~\cite{chi2023diffusionpolicy}, building on the foundations of denoising diffusion probabilistic models~\cite{NEURIPS2020_4c5bcfec} and score-based generative modeling~\cite{pmlr-v37-sohl-dickstein15}, formulates a robot’s visuomotor policy as a conditional denoising diffusion process over the action space. Rather than directly regressing an action, the policy learns the score (gradient) of the action distribution conditioned on observations and refines noise into actions through stochastic Langevin dynamics during inference.

This approach brings several key benefits:
    \begin{itemize}
        \item It naturally captures multimodal action distributions, a common challenge in robotic policy learning.
        \item It gracefully scales to high-dimensional action spaces, enabling the prediction of sequences of future actions for better temporal consistency.
        \item It ensures training stability, avoiding the instability of energy-based models by learning the score directly and sidestepping the need to estimate intractable normalization constants.
    \end{itemize} 

To make diffusion policies practical on physical robots, it was introduced a Receding-horizon control, enabling closed-loop action execution with continuous re-planning, visual conditioning, processing observations once to condition all denoising steps efficiently and a time-series diffusion transformer, reducing over-smoothing and improving performance in tasks requiring rapid, precise actions.

It was tested across 12 different tasks spanning four manipulation benchmarks (simulated and real-world), Diffusion Policy achieves a consistent performance boost of around 46.9\% average improvement over state-of-the-art methods.


\section{Methodology}

This work builds upon the GenDP framework~\cite{wang2023gendp}, which integrates Diffusion Policy-based visuomotor learning with 3D semantic field representations (via D³Fields~\cite{wang2024d3fields} ) for category-level generalization in robotic manipulation. The original GenDP implementation targets a bimanual ALOHA robot platform equipped with four Intel RealSense cameras, using a ROS-based control interface for both perception and actuation. While effective in its original form, the framework was designed for a fixed hardware configuration, limiting its applicability to other robotic platforms and sensor modalities.

\begin{figure}

\centering
\subfloat[Experimental setup of the UR5 robot with four Azure Kinect cameras.]{\label{fig:ur5setup}
\centering
\includegraphics[width=0.45\linewidth]{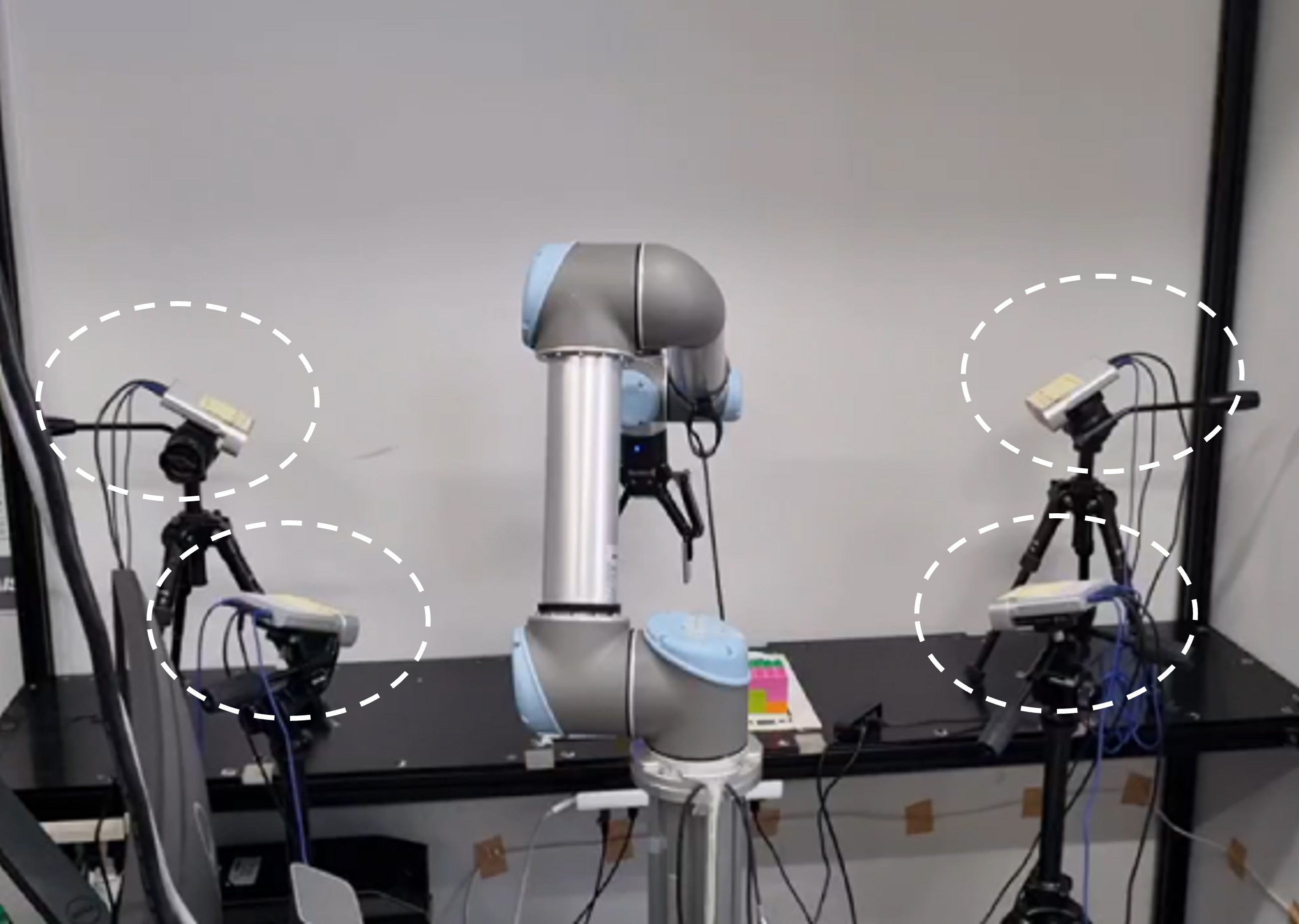}
}
\hfill
\subfloat[Experimental setup of the ALOHA system with three Intel RealSense cameras.]{\label{fig:alohasetup}
\centering

\includegraphics[width=0.45\linewidth]{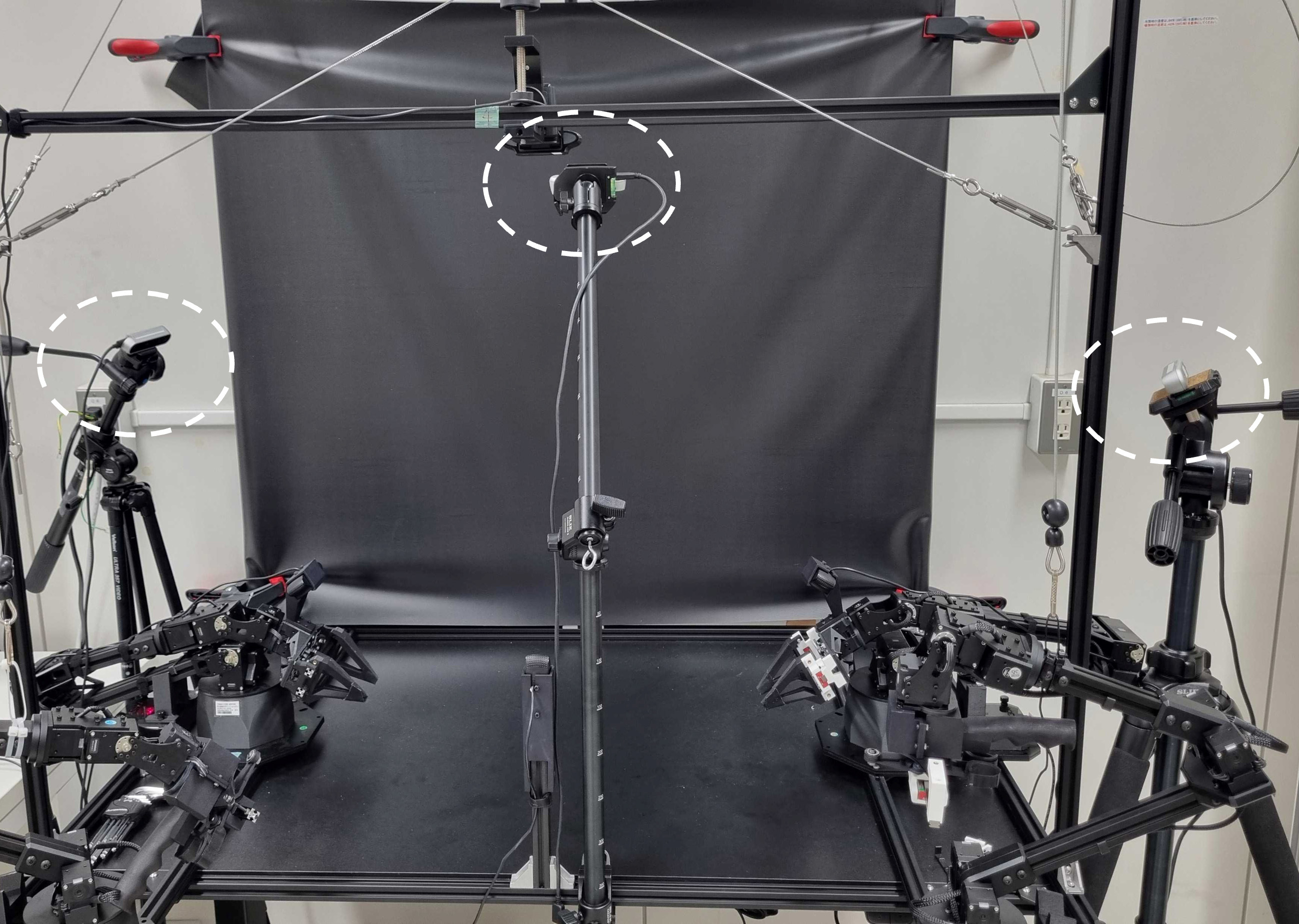}
}

\caption{Sample setups used in our experiments.}
\label{fig:csetup}
\end{figure}

To overcome these limitations, we developed a modified and extended framework that enables the seamless integration of alternative robot–camera setups. Specifically, we replaced the ALOHA bimanual manipulator and RealSense depth cameras with a UR5 robotic arm controlled via a SpaceMouse 3D input device, paired with four Microsoft Azure Kinect cameras, as shown in Fig. \ref{fig:ur5setup}. In addition, we implemented a configuration layer that allows users to switch between the original ALOHA + RealSense pipeline and the new UR5 + Azure Kinect pipeline with minimal changes, supporting data collection, training, and evaluation across both workcells.

Beyond this specific migration, the overarching methodology is to establish a flexible framework in which perception, control, and learning components are separated from the underlying robot and sensor hardware. By abstracting hardware dependencies, the workcell can be extended to support additional robotic manipulators and sensor configurations without major code rewrites. This adaptability aims to accelerate research by allowing direct comparisons across varied platforms, encourage reproducibility, and enabling broader adoption of field-based visuomotor learning methods in diverse settings.

\section{Implementation}

\begin{figure*}[t]

\centering
\subfloat[General structure of a workcell, consisting of a robot, a control interface, and a camera system.]{\label{fig:generalworkcell}
\centering
\includegraphics[width=0.4\linewidth]{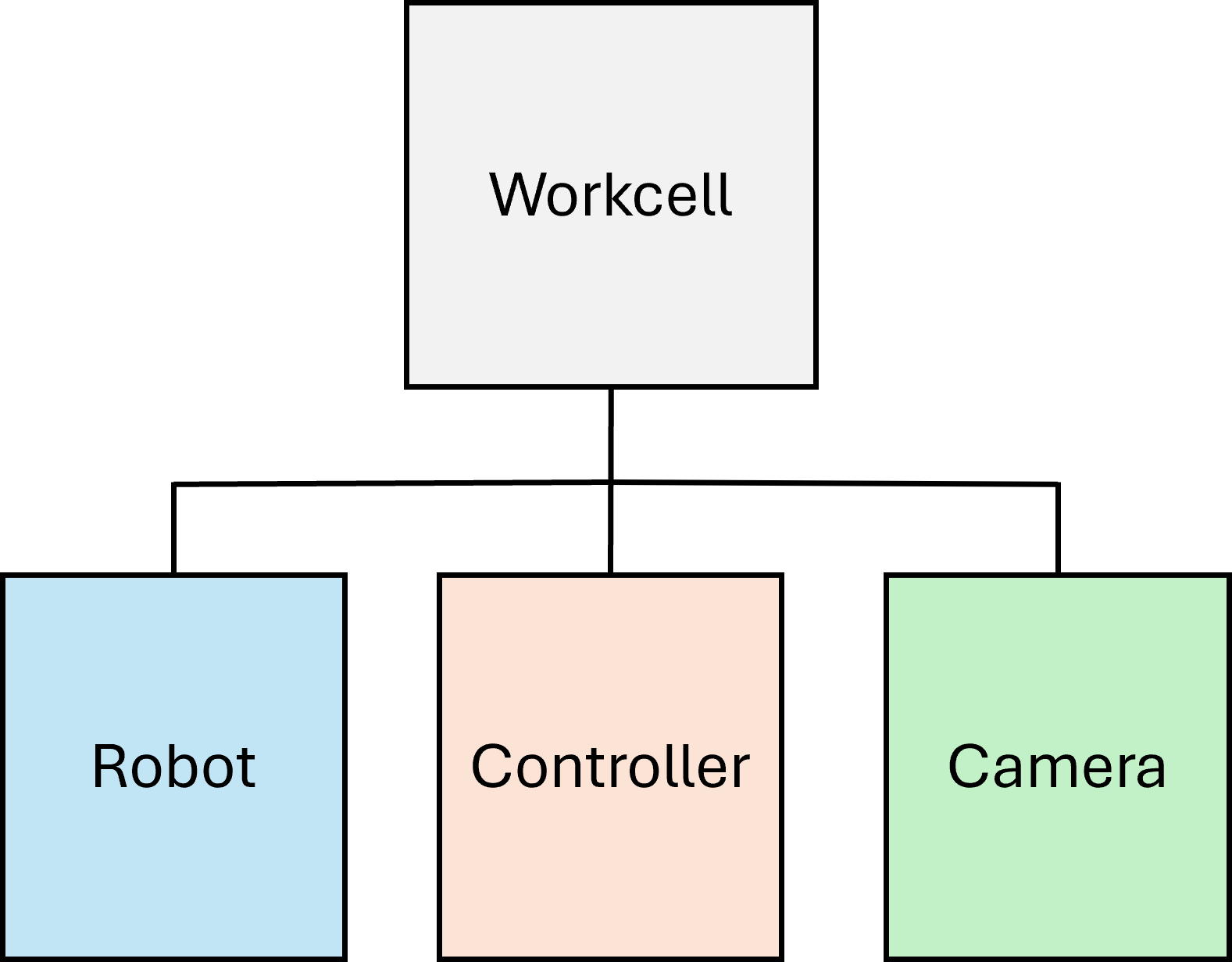}
}
\hfill
\subfloat[Examples of specific workcells: one with a UR5 manipulator, a SpaceMouse controller, and Azure Kinect cameras; another with the ALOHA bimanual platform, its dual-hand controller, and Intel RealSense cameras. The ellipsis indicates that additional workcell configurations can be supported.]{\label{fig:typeworkcell}
\centering

\includegraphics[width=0.4\linewidth]{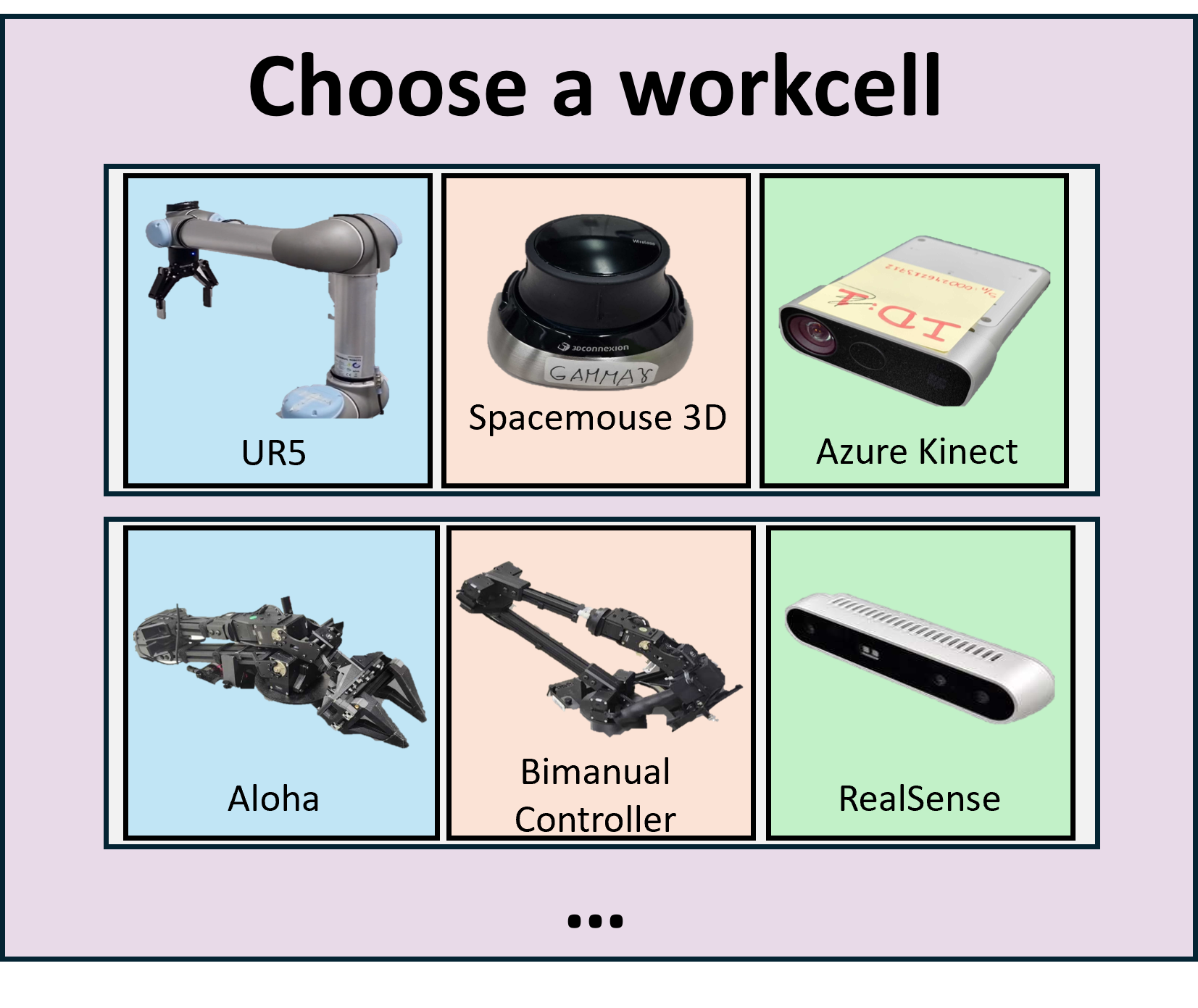}
}

\caption{Modular workcell concept. (a) A workcell is defined as a combination of a robot, a control interface, and a camera system. (b) Examples of supported workcells, illustrating the flexibility of the framework to integrate different robots, controllers, and sensors.}
\label{fig:setup}
\end{figure*}

The UR5 robotic manipulator was integrated using the Real-Time Data Exchange (RTDE) interface provided by Universal Robots. This allowed low-latency and direct control of the manipulator’s joint states, enabling real-time teleoperation through a SpaceMouse controller. The SpaceMouse provides six degrees of freedom (translation and rotation), which were mapped to Cartesian velocity commands for the UR5’s end-effector.
For the ALOHA platform, originally implemented in ROS 1, we ported the control and communication stack to ROS 2 to ensure compatibility with modern middleware and to unify the interface with our updated workcell (Fig. \ref{fig:setup}). This allowed both robot types to be operated and monitored under a consistent software structure, facilitating the switch between platforms without altering core learning or perception modules.

To develop the custom scripts for controlling the UR5 via the SpaceMouse without ROS, we drew inspiration from the open-source Gello Software~\cite{10801581} repository, which already implements SpaceMouse teleoperation for the UR5 through the RTDE interface. This provided a proven foundation for non-ROS, low-latency control, allowing us to adapt and extend their approach to fit our episodic data collection framework and multi-camera integration.

The original GenDP pipeline is tightly coupled with Intel RealSense hardware and corresponding SDKs. To enable compatibility with the Azure Kinect cameras, we designed a new acquisition module that captures synchronized RGB-D frames from four Azure Kinect devices in real time. The synchronization relies on hardware triggering, with one device configured as the master and the remaining three as subordinates, following the multi-camera setup recommended by Microsoft for time alignment. This ensures that all RGB-D streams are temporally consistent with the robot motion data.
The captured frames are passed through the D³Fields feature extraction pipeline, which was modified to replace the RealSense-specific intrinsic and extrinsic calibration routines with Azure Kinect calibration parameters. We further optimized the multi-threaded capture process to ensure low-latency data storage, enabling the collection of high-quality episodic datasets without dropped frames.

A core contribution of this work is a custom-built episodic recording system that simultaneously logs robot state data and synchronized multi-camera RGB-D streams. For the UR5 setup, this includes joint positions, velocities, and end-effector poses obtained via RTDE, sampled at the same frequency as the visual data. For the ALOHA setup, equivalent state data is collected through the ROS 2 interface.

The recording module supports flexible task definition: during teleoperation with SpaceMouse (for UR5) or ALOHA leader–follower controllers, operators can perform arbitrary manipulation trajectories, which are recorded as “episodes” along with all associated sensory data. These episodes form the training data set for subsequent policy learning. The system ensures that data is stored in a format directly compatible with the GenDP training scripts.

Following data collection, we trained visuomotor policies using the Diffusion Policy architecture provided in the original GenDP codebase. The collected episodes were processed through the adapted D³Fields pipeline, generating 3D semantic descriptors from the Azure Kinect streams. These features were combined with raw point clouds and used as input to the PointNet++~\cite{NIPS2017_d8bf84be} backbone within the GenDP policy model.
We developed a new evaluation module based on GenDP’s original test scripts, modified to interface with both the UR5 and ALOHA control stacks. This allowed for systematic benchmarking of trained policies on the same manipulation tasks used in data collection. 

\begin{figure}
    \centering
    \includegraphics[width=0.3\linewidth]{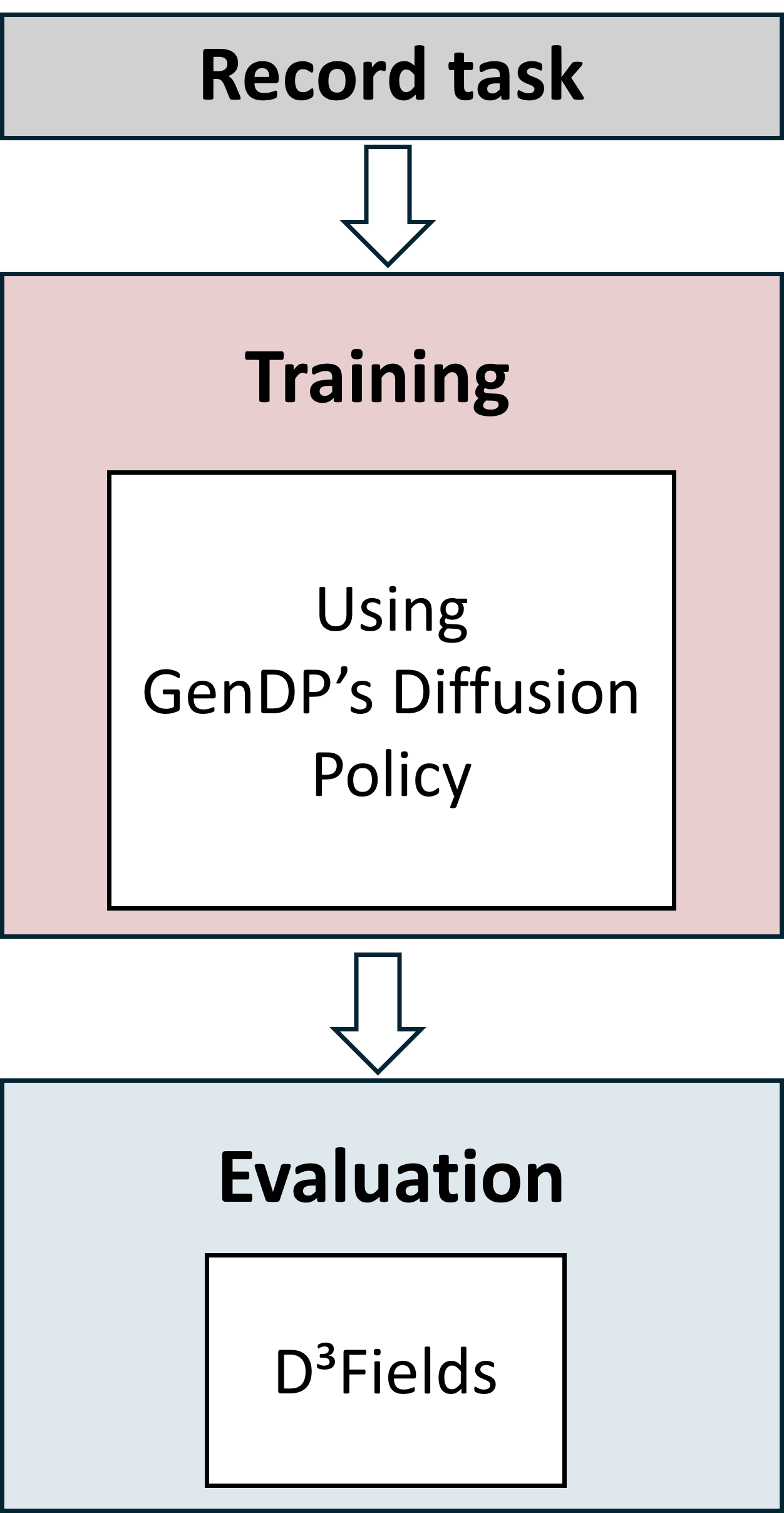} 
    \caption{Workflow of the proposed framework. Episodic demonstrations of a manipulation task are recorded, a visuomotor policy is trained using the GenDP pipeline, and the learned policy is evaluated on the chosen hardware setup using the D³Fields feature extraction pipeline..}
    \label{fig:workflow} 
\end{figure}

To maximize reusability and scalability, we implemented a modular configuration layer in the codebase (Fig.~\ref{fig:workflow}). This layer allows the operator to select between the UR5 + Azure Kinect or ALOHA + RealSense workcells at runtime. Switching involves only a configuration file change, automatically adjusting device initialization and calibration loading to match the chosen hardware. This design enables researchers to deploy the framework in different labs and hardware configurations without significant code rewriting.

The overall goal of these modifications is to support robot learning through randomization of existing scenes using geometry-preserving transformations, while providing the flexibility to interchange hardware configurations. By allowing the same perception–control–learning loop to operate in different robot–camera setups, we facilitate comparative studies, improve generalization analysis, and expand the applicability of diffusion policy–based visuomotor learning frameworks beyond their original hardware constraints.


\section{Evaluation}

\begin{figure}

\centering
\subfloat[Grasp-and-lift task performed by an UR5 robot with an Robotiq 2F-140 gripper.]{\label{fig:ur5task}
\centering
\includegraphics[width=0.45\linewidth]{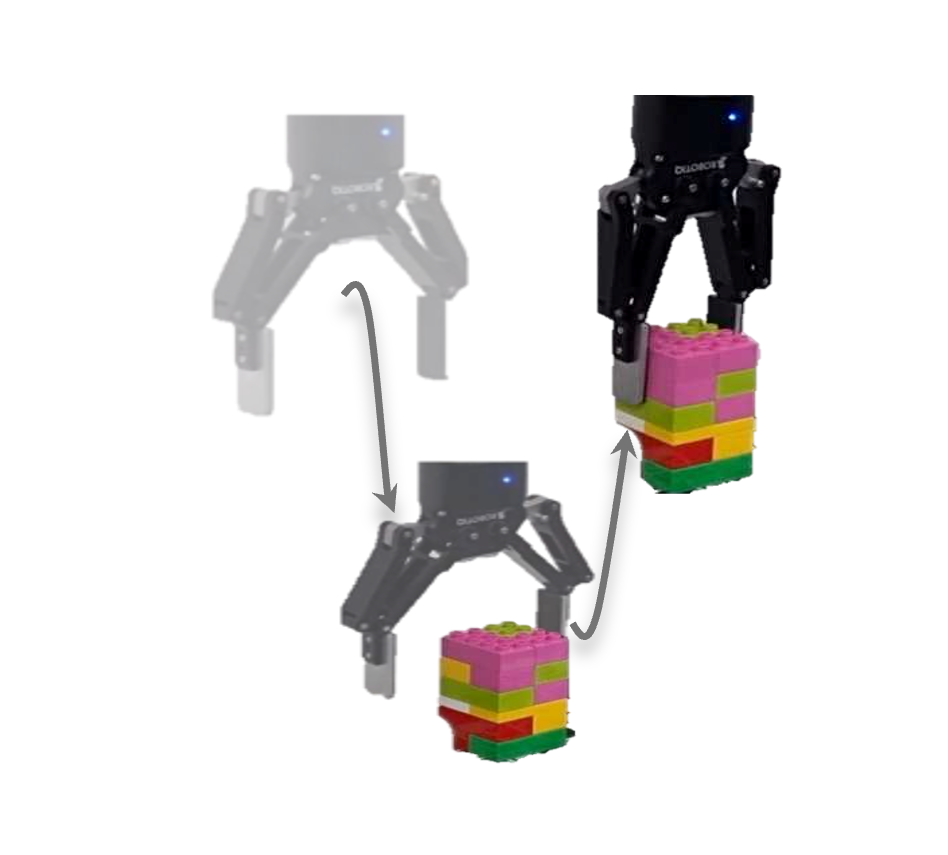}
}
~
~
\subfloat[Grasp-and-lift task performed by an ALOHA robot with the standard ALOHA gripper.]{\label{fig:alohatask}
\centering

\includegraphics[width=0.45\linewidth]{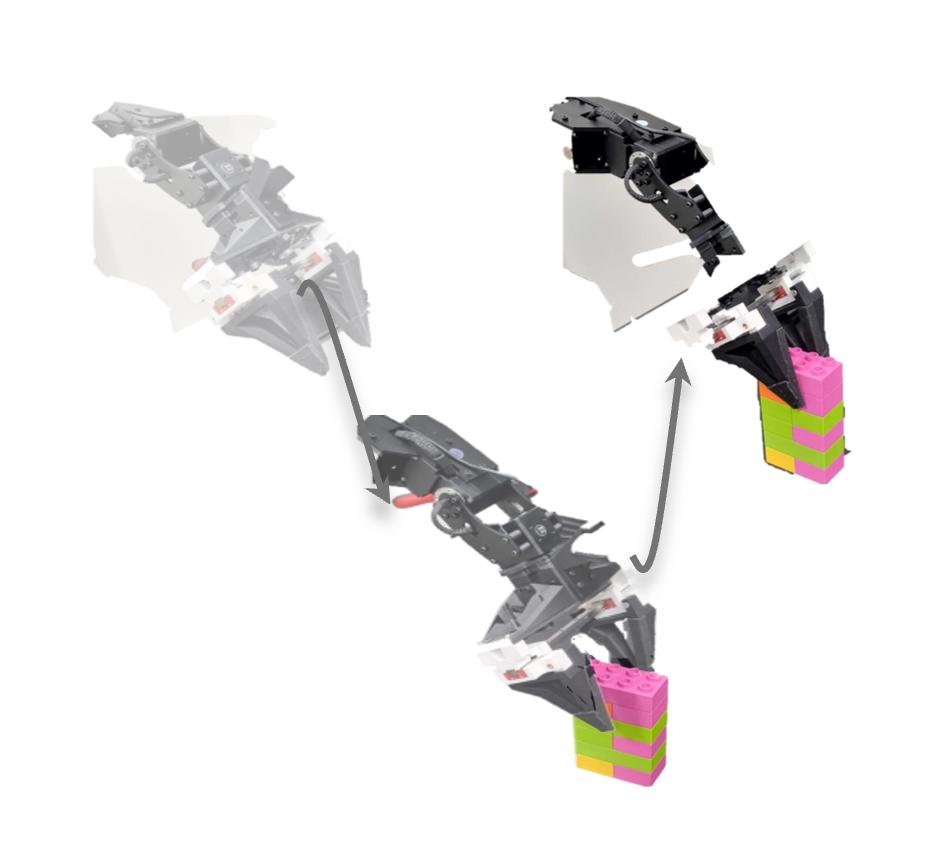}
}

\caption{Example of the grasp-and-lift task: the selected robot approaches the target object, grasps it, and lifts it from the table.}
\label{fig:csetup}
\end{figure}

The evaluation focused on a single object manipulation task involving a block made of LEGO/DUPLO-style pieces as illustrated in Fig.~\ref{fig:ur5task} and ~\ref{fig:alohatask}. In each episode, the robot was initialized from a predefined starting configuration and tasked with locating the target block, grasping it, and lifting it a few centimeters above the table surface. The object’s position was varied slightly between episodes to introduce minor spatial variability while maintaining similar overall scene configurations.
A total of 100 episodes were collected using the UR5 + Azure Kinect workcell, with each episode lasting approximately 15 seconds. Additionally, we conducted a separate set of experiments on the Aloha robot equipped with Intel RealSense cameras, following the same task definition and evaluation protocol. During data collection, we recorded:
    \begin{itemize}
        \item Robot kinematics: joint positions and end-effector poses at each timestep.
        \item Visual data: synchronized RGB and depth images from all four Azure Kinect cameras (UR5 setup) or RealSense cameras (Aloha setup).
        \item Calibration and extrinsics: intrinsic parameters and spatial transformations for multi-camera alignment.
    \end{itemize} 
 This data set was subsequently used to train the policy model.

The training phase employed the Diffusion Policy model implementation provided in GenDP, adapted for the UR5 + Azure Kinect configuration as well as for the Aloha + RealSense configuration. The collected episodes were processed through the modified D³Fields pipeline to generate 3D semantic descriptors from multi-view RGB-D data, which were combined with point cloud features as policy inputs. The model was trained for 8,000 epochs with a batch size of 64 and 8 parallel data loader workers, using a high-performance workstation equipped with an NVIDIA GPU. The optimizer and learning schedule followed the original GenDP training defaults, ensuring consistency with prior results while adapting only the perception and control interfaces.

After training, we conducted an evaluation of 20 rollouts using the learned policy in the same physical setup as the training data. Each rollout followed the same task definition as the recorded episodes: the robot had to autonomously move from its initial pose, identify the block, grasp it, and lift it off the table. 
For evaluation, we selected model checkpoints corresponding to the 200–400 epoch range, as preliminary training curves indicated that this interval consistently produced the best performance in terms of validation metrics. During each evaluation trial, task success was defined as completing the full grasp-and-lift sequence without dropping the object or colliding with the environment.

The trained policy achieved a success rate of 80\% over 20 evaluation trials in the UR5 + Azure Kinect setup. When evaluated on the Aloha + RealSense workcell, the same framework achieved a higher success rate of 90\%, highlighting its ability to transfer effectively across different robot–camera configurations. The failure cases in both setups typically involved either imprecise grasping due to slight perception errors in block localization, or premature release of the object during the lifting phase. These failure modes suggest that further improvements may be achieved by incorporating additional demonstrations with greater object position diversity, enhancing grasp robustness, or fine-tuning the perception model for small objects.

Overall, the evaluation confirms that the modified framework retains the strong learning capabilities of the original GenDP system while extending its applicability to different robot–camera configurations. The higher success rate observed in the Aloha + RealSense setup further underscores the robustness of the approach. The ability to achieve high success rates after only 100 demonstration episodes emphasizes the data efficiency of diffusion policy–based visuomotor learning when combined with rich 3D semantic scene representations. This experiment also highlights the importance of checkpoint selection in diffusion policy training: rather than simply using the final epoch, monitoring training curves and selecting the most promising intermediate models can yield better real-world performance, as observed in the 200–400 epoch range.


\section{Conclusion and Discussion}
This work presented an extended version of the GenDP framework, adapting it from its original ALOHA + RealSense configuration to operate seamlessly with a UR5 + Azure Kinect workcell while preserving compatibility with the original hardware setup. By redesigning the control, perception, and data acquisition modules, we enabled robust policy training and evaluation across different robot–camera combinations without altering the core learning architecture. Our evaluation on a grasp-and-lift manipulation task involving a LEGO/DUPLO block demonstrated that the adapted framework can achieve high success rates across both platforms: 80\% with the UR5 + Azure Kinect setup and 90\% with the ALOHA + RealSense setup.

These results validate the feasibility of transferring a diffusion policy–based visuomotor learning pipeline to new hardware configurations with minimal adjustments. The system maintained strong generalization capabilities despite modest scene variability, highlighting the benefits of combining diffusion-based action generation with rich 3D semantic scene representations via the D³Fields pipeline. Notably, the two platforms differed in the action representation used during training (joint-space actions for UR5 versus end-effector actions for ALOHA, later mapped to joints for execution), which may have influenced the observed outcomes. We emphasize, however, that the primary contribution of this work is not to contrast performance across platforms, but to demonstrate that the same learning framework can be effectively deployed in distinct robot–camera configurations with consistently strong results.

\subsection{Future work}
Several avenues for further development emerge from this research. We plan to extend the system’s multi-platform flexibility by enabling full interchangeability between robots and camera systems, regardless of their original hardware pairing. This would allow configurations such as UR5 + RealSense, ALOHA + Azure Kinect, or new combinations involving other manipulators such as the Franka Emika Panda, paired with a variety of depth-sensing technologies. To achieve this, we will expand the framework’s hardware abstraction layer to support a broader spectrum of perception and control modalities, including additional RGB-D sensors with varying resolution, depth and latency characteristics and integrating ROS2 or different robot's APIs.

Finally, future work will explore multi-robot and heterogeneous setups, in which multiple manipulators equipped with different sensors operate within the same environment and share a unified policy learning pipeline. By pursuing these directions, we expect to further expand the applicability of the proposed framework, enabling more realistic, diverse, and challenging manipulation scenarios that better reflect the complexities of real-world environments.


\section{Acknowledgement}
This work was carried out with financial support and access to the experimental facilities of the National Institute of Advanced Industrial Science and Technology (AIST). We sincerely appreciate the valuable assistance and guidance received throughout the course of this research. We would also like to thank Andrea M. Salcedo-Vazquez, Olivier Crampette, Muhammad A. Muttaqien, and Abdullah Mustafa for their support and insightful discussions.

\addtolength{\textheight}{-12cm}   


\flushend
\bibliographystyle{IEEEtran} 
\bibliography{IEEEabrv, references}

\end{document}